\documentclass[letterpaper]{article} 
\usepackage{aaai24} 
\usepackage{times}  
\usepackage{helvet}  
\usepackage{courier}  
\usepackage[hyphens]{url}  
\usepackage{graphicx} 
\usepackage{natbib}  
\usepackage{caption} 
\usepackage[T1]{fontenc}
\usepackage{url}
\usepackage{booktabs}
\usepackage{pifont}
\usepackage{colortbl}
\definecolor{mygray}{gray}{0.6}
\newcommand{\algcom}[1]{\textsl{\color{mygray}{\footnotesize #1}}}
\usepackage[linesnumbered,ruled]{algorithm2e}
\usepackage{subfigure}
\usepackage{amsthm}
\usepackage{enumerate}
\usepackage{multirow}
\usepackage{amssymb}
\usepackage{float}
\usepackage{array}
\usepackage{amsmath,amsfonts}
\usepackage{bm,dsfont}

\def\eqref#1{equation~\ref{#1}}

\def\1{\bm{1}}

\DeclareMathAlphabet{\mathsfit}{\encodingdefault}{\sfdefault}{m}{sl}
\SetMathAlphabet{\mathsfit}{bold}{\encodingdefault}{\sfdefault}{bx}{n}

\newcommand{\PP}{\mathbb P}

\usepackage{tcolorbox}
\definecolor{grey}{rgb}{0.33, 0.33, 0.33}

\newcommand{\squishlist}{
\begin{list}{{{\small{$\bullet$}}}}
{\setlength{\itemsep}{1pt}      \setlength{\parsep}{0pt}
\setlength{\topsep}{-2pt}       \setlength{\partopsep}{0pt}
\setlength{\leftmargin}{1em} \setlength{\labelwidth}{1em}
\setlength{\labelsep}{0.5em} } }
\newcommand{\squishend}{  \end{list}  }

\makeatletter
\renewcommand*\env@matrix[1][*\c@MaxMatrixCols c]{%
  \hskip -\arraycolsep
  \let\@ifnextchar\new@ifnextchar
  \array{#1}}
\makeatother

\usepackage{textcomp}
\usepackage{verbatim}
\usepackage{graphicx}

\newcommand{\fedual}{FedFixer}
\newtheorem*{defn}{Definition}
\newtheorem{theorem}{Theorem}

\ifodd 0
\newcommand{\yl}[1]{\textbf{\color{red}(Yang: #1)}}
\newcommand{\zzw}[2]{\textbf{\color{blue}(Zhaowei: #1)}{\color{blue}~#2}}
\newcommand{\xy}[1]{\textbf{\color{brown}(Xinyuan: #1)}}
\newcommand{\olga}[1]{\textbf{\color{black}(Olga: #1)}}
\else
\newcommand{\yl}[1]{}
\newcommand{\zzw}[2]{}
\newcommand{\xy}[1]{}
\newcommand{\olga}[1]{}
\fi

\usepackage{newfloat}
\usepackage{listings}
\DeclareCaptionStyle{ruled}{labelfont=normalfont,labelsep=colon,strut=off} 
\floatstyle{ruled}
\newfloat{listing}{tb}{lst}{}
\floatname{listing}{Listing}
\title{\fedual: Mitigating Heterogeneous Label Noise in Federated Learning}

\author{ Xinyuan Ji\textsuperscript{\rm 1, \rm 2}, 
         Zhaowei Zhu\textsuperscript{\rm 3}, 
         Wei Xi\textsuperscript{\rm 1}\equalcontrib,
         Olga Gadyatskaya\textsuperscript{\rm 2},
         Zilong Song\textsuperscript{\rm 1},
         Yong Cai\textsuperscript{\rm 4},
         Yang Liu\textsuperscript{\rm 5}\equalcontrib
         }

\affiliations{
\textsuperscript{\rm 1}Xi'an Jiaotong University\\
\textsuperscript{\rm 2}Leiden University\\
\textsuperscript{\rm 3}Docta.ai\\
\textsuperscript{\rm 4}IQVIA Inc \& California State University, Monterey Bay\\
\textsuperscript{\rm 5}University of California, Santa Cruz\\ 
\{jixinyuan1996, weixi.cs\}@gmail.com, yangliu@ucsc.edu
}

\begin{document}
\maketitle
\begin{abstract}
Federated Learning (FL) heavily depends on label quality for its performance.
{\color{black}However, the label distribution among individual clients is always both noisy and heterogeneous. }
The high loss incurred by client-specific samples in heterogeneous label noise poses challenges for distinguishing between client-specific and noisy label samples, impacting the effectiveness of existing label noise learning approaches. 
{\color{black}To tackle this issue, we propose \fedual, where the personalized model is introduced to cooperate with the global model to effectively select clean client-specific samples.}
{\color{black}In the dual models, updating the personalized model solely at a local level can lead to overfitting on noisy data due to limited samples, consequently affecting both the local and global models' performance.}
To mitigate overfitting, we address this concern from two perspectives. Firstly, we employ a confidence regularizer to alleviate the impact of unconfident predictions caused by label noise. Secondly, a distance regularizer is implemented to constrain the disparity between the personalized and global models.
We validate the effectiveness of \fedual\ through extensive experiments on benchmark datasets. The results demonstrate that \fedual\ can perform well in filtering noisy label samples on different clients, especially in highly heterogeneous label noise scenarios.

\end{abstract}

\section{Introduction}
\label{intro}

Federated Learning (FL) aims to learn a common model from different clients while maintaining client data privacy, and it has gradually been applied to real-world applications \citep{li2020federated,he2020fedml,he2019central,pu2023federated}. However, the presence of heterogeneous {\color{black}label noise} \cite{li2019convergence,li2021fedbn, zhu2021federated} in each local client severely degrades the generalization performance of FL models \citep{wang2022fednoil}. 

\begin{figure*}[t]
\centering
\subfigure[Label noise distribution on client $p$ and $q$. Black are correct labels and Red are noisy labels. ]{
\begin{minipage}{11.2cm}
\centering
\includegraphics[width = 1 \textwidth]{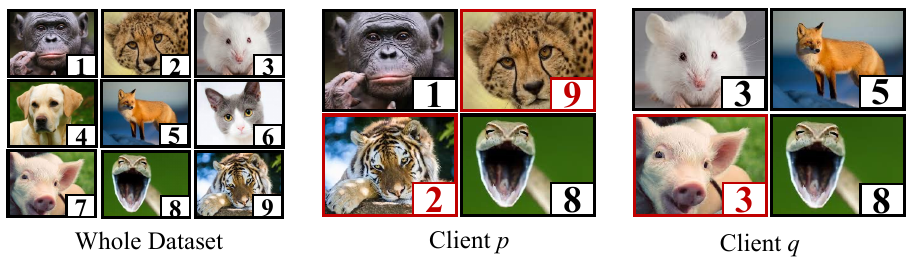}
\label{intro_excample}
\end{minipage}
}
\subfigure[CL with noisy labels \& FL with heterogeneous noisy labels.]{
\begin{minipage}{6.1cm}
\centering
 \includegraphics[width = 1 \textwidth]{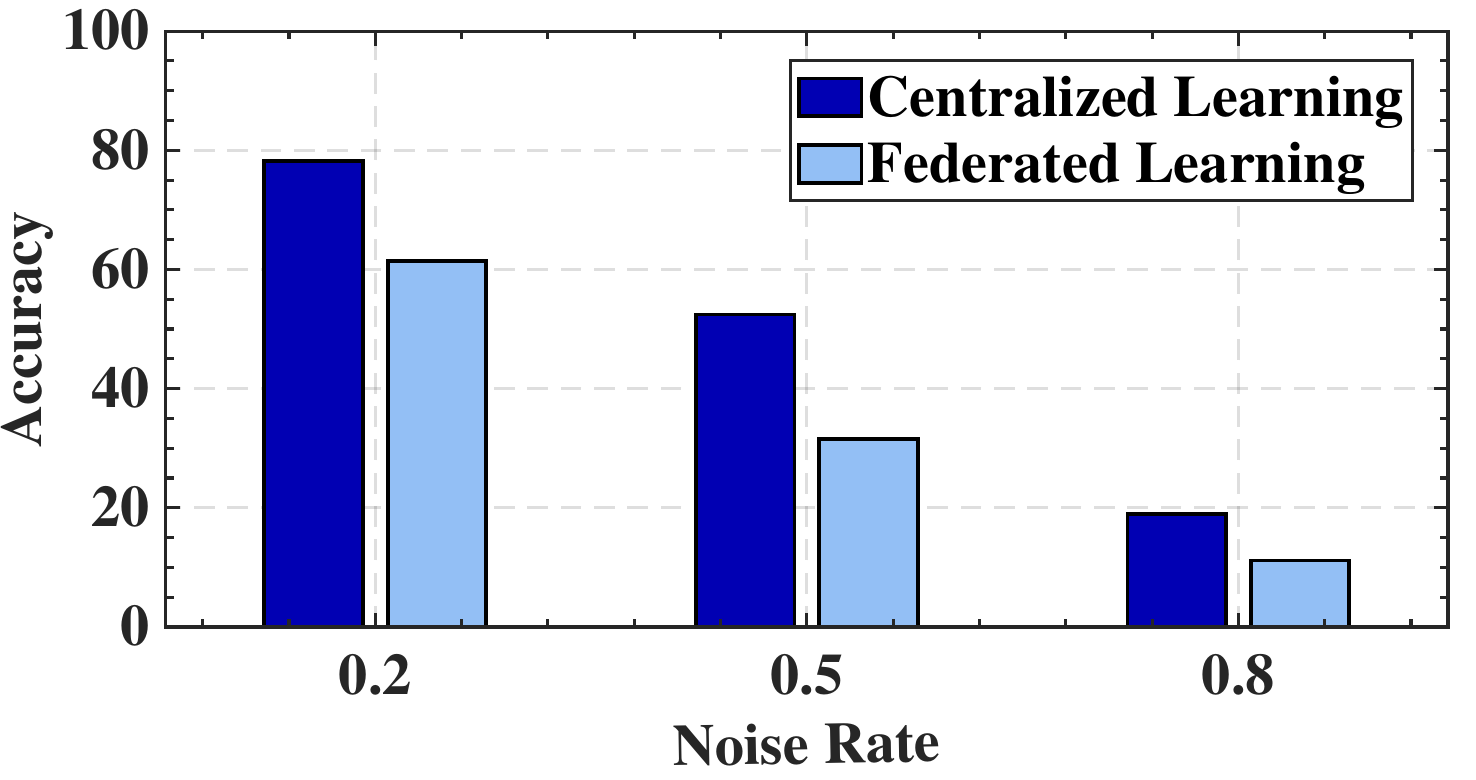}
\label{intro_comparation}
\end{minipage}
}
\caption{(a): The heterogeneous label noise distributions encompass diverse true class samples or varying label noise transitions. (b): The performance comparison between CL and FL on CIFAR-10 with the rate of label noise 0.2, 0.5, and 0.8.}
\label{fig:figure1}
\end{figure*}

Unlike Centralized Learning (CL) with noisy labels \cite{zhu2023unmasking, zhu2022beyond, zhu2022detecting}, FL confronts a unique situation where each client's label noise distribution\footnotemark may be heterogeneous due to variations in the {\color{black}true class distribution} and personalized human labeling errors \cite{wei2022learning,han2018co,yi2022learning}. Fig. \ref{intro_excample} visually {\color{black}demonstrates significant differences in the label noise distribution} between client $p$ and client $q$. Consequently, the existence of {\color{black}heterogeneous label noise} severely degrades the generalization performance of FL model \cite{wang2022fednoil}, leading to an even more pronounced impact on FL compared to CL with noisy labels, as shown in Fig. \ref{intro_comparation}. Therefore, there is a critical need for a robust label noise learning method tailored to FL to address this challenge effectively. 

Numerous research efforts have been devoted to addressing noisy labels in CL \cite{sieve2020,li2019dividemix,liu2020peer,natarajan2013learning,wei2020combating}. However, when directly applied to FL, these CL-based methods encounter significant challenges. For instance, learning a model using local noisy label data on the client side is susceptible to overfitting due to the limited sample size. Furthermore, training a global model through FedAvg \cite{mcmahan2017communication} struggles to effectively learn client-specific samples, potentially leading to difficulties in correctly identifying whether high-loss samples are noisy label samples or client-specific samples. Moreover, existing methods designed for the issue of label noise in FL can be broadly categorized into coarse-grained \cite{chen2020focus,li2021federated,yang2021client} and fine-grained \cite{tuor2021overcoming,xu2022fedcorr} methods. However, these methods usually ignore the heterogeneity of noisy label samples in FL. They encounter difficulty in the personalized identification of noisy label samples across diverse clients. This difficulty impairs their potential to achieve substantial performance improvements.
It is challenging to effectively recognize wrongly labeled samples from different clients in FL. 

In this paper, we propose \fedual, an FL algorithm with the dual model structure to mitigate the impact of heterogeneous noisy label samples in FL. 
\fedual\ consists of two models: the global model ($w$) and the auxiliary personalized model ($\theta$). During the local training process at each client, the global model $w$ and personalized model $\theta$ are alternately updated using the samples selected by each other. {\color{black}With alternative updates of dual models, \fedual\ can effectively decrease the risk of error accumulation from a single model over time.}
Unfortunately, {\color{black}unlike the global model which benefits from updates through aggregation, the local updates of the personalized model are prone to overfitting on noisy label data due to being confined to a limited sample size.} This susceptibility to overfitting has negative implications on the overall performance of the dual model structure. 
To alleviate the overfitting in the dual models, it can be addressed from two different perspectives.
First, we employ a confidence regularizer to alleviate the unconfident predictions induced by label noise. Second, we implement a distance regularizer between the parameters of the personalized model and the global model to counter the risk of the personalized denoising model overfitting local data. 

Our contributions can be summarized as follows: 
\begin{enumerate}[(1)]
\item \textbf{Dual Model Structure}: Our tailored dual model structure for FL with heterogeneous label noise effectively adapts to heterogeneous label noise distributions, addressing challenges from varying {\color{black}true class distributions} and personalized human labeling errors. 

\item \textbf{Regularization Terms}: To combat the overfitting of dual models, we first employ the confidence regularizer to restrict unconfident predictions caused by label noise. Secondly, we implement a distance regularizer into the dual model structure to constrain model distance with the global model.

\item \textbf{Extensive Experimental Validation}: Validated on the benchmark datasets with multiple clients and varying degrees of label noise, \fedual\ demonstrates comparable performance to state-of-the-art (SOTA) methods in general scenarios with heterogeneous noisy labels. However, in highly heterogeneous label noise scenarios, our method outperforms existing SOTA methods by up to 10\%, showcasing its effectiveness in addressing heterogeneous label noise in FL.
\end{enumerate}

\begin{figure*}[t]
\centering
\begin{center}
    \includegraphics[height=4.6cm]{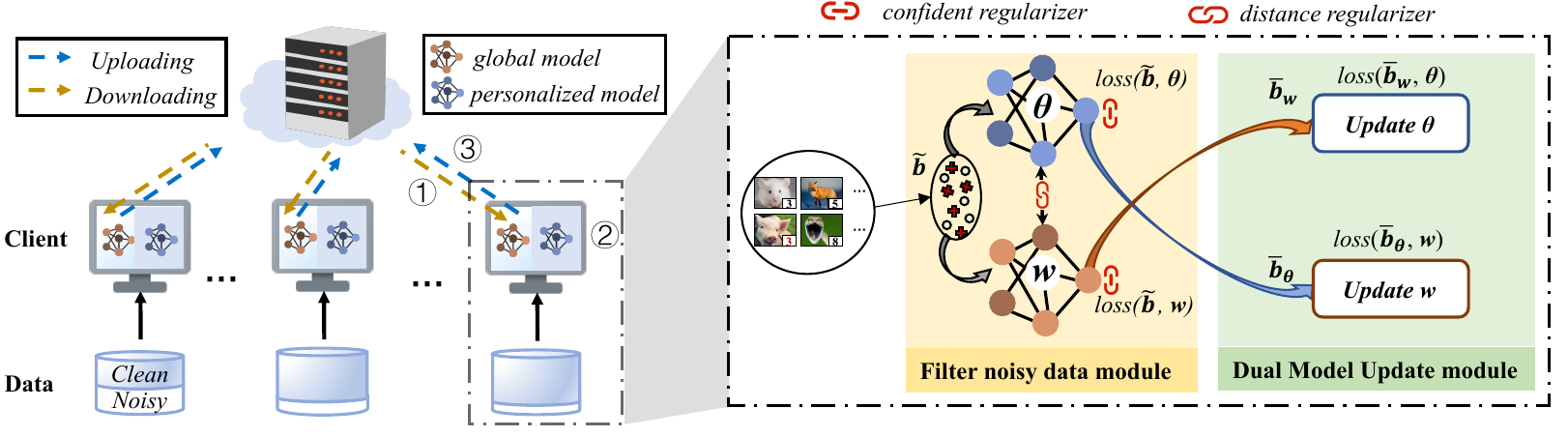}
    \caption{Overall architecture. The training process of the proposed \fedual\ has three stages: \ding{172} deployment of the global model, \ding{173} the local model updates, and \ding{174} the global model aggregation. In the second stage, dual models are alternately updated (in the Dual Model Update module) based on the selected samples (in the Filter Noisy Data module) by each other.}
    \label{fig:figure2}
\end{center}
\end{figure*}

\section{Related Work}
\label{related work}

In this section, we review the related works. We first discuss existing robust FL methods that address label noise. Additionally, we introduce the related work of learning with noisy labels in a centralized setting.

\paragraph{Robust FL} 
Robust FL has been extensively studied, and it can be categorized into client-level and sample-level methods. For client-level approaches, Yang et al. \cite{yang2021client} proposed an online client selection framework that efficiently selects a client set with low noise ratios using Copeland score and multi-arm bandits. Similarly, Fang et al. \cite{fang2022robust} provided the RHFL to address FL problem with noisy and heterogeneous clients. On the other hand, sample-level methods focus on reducing noisy label samples from local data. Tuor et al. \cite{tuor2021overcoming} proposed a method for the distributed method to select relevant samples, where the high-relevance data samples at each client are identified using a benchmark model trained on a small benchmark dataset, improving model's generalization ability \citep{pu2021lifelong,pu2022meta,pu2023memorizing}. Xu et al. \cite{xu2022fedcorr} proposed FedCorr, {\color{black}a general multi-stage framework. The initial step involves the detection of clean client sets using GMM(Gaussian Mixture Models)~\citep{pu2020dual}. Subsequently, model fine-tuning is performed on these clean client sets, and the corrected label samples are relabeled. Finally, the federated model is trained using the conventional FL stage.} However, these existing methods can't effectively address the challenge of heterogeneous label noise in FL. In contrast, our work proposes a dual model structure approach to effectively adapt to heterogeneous label noise distributions to address this challenge.

\paragraph{Learning with Noisy Labels} 

Learning with noisy labels is a well-studied area, primarily focused on centralized learning \cite{zhu2021second, zhu2021clusterability, cheng2023mitigating, wei2022aggregate}. Existing approaches can be categorized into several categories, including Robust Architecture \cite{goldberger2016training,lee2019robust}, Robust Regularization~\cite{xia2020robust,gudovskiy2021autodo}, Robust Loss Design~\citep{ghosh2017robust,pu2023dynamic,jiang2021information}, and Sample Selection~\cite{jiang2018mentornet, han2018co, sieve2020}.
In this paper, we specifically review Sample Selection methods, which have gained attention due to the theoretical and empirical exploration of DNNs' memorization properties to identify clean examples from noisy training data \cite{xia2020robust,liu2020early}. For instance, Han et al. proposed the ``Co-teaching'' deep learning paradigm to combat noisy labels, while Cheng et al. proposed CORES$^2$, which progressively sieves out corrupted examples. Additionally, Xia et al. \cite{xia2020robust} proposed robust early-learning to reduce the side effects of noisy labels before early stopping and enhance the memorization of clean labels. While existing research has achieved promising results in noisy label learning within the centralized learning framework, direct application of these approaches to FL is challenging. This is due to the limited sample size and the heterogeneous label noise distribution, which can result in uncertain performance for each client. Therefore, our proposed method specifically addresses these challenges to effectively handle heterogeneous label noise in FL.

\section{Problem Definition}
\label{sec:3}
 
In this section, we formulate the problem of FL with heterogeneous label noise. Consider a large collection of data $D$ with a sample size of $N$ and $L$ class labels, which is distributed over $K$ clients that $D = \cup_{k=1}^{K}D^{k} := \cup_{k=1}^{K}\left\{(x^k_{n},y^k_{n})\right\}_{n \in [n_{k}]}$, where $[n_{k}] = \left\{1, 2, ..., n_{k}\right\}$ is the set of example indices on client $k$. Each local instance $(x^k_n,y^k_n)\in D^k$ follows the data distribution $(X^k,Y^k)\sim\mathcal D^k$. If the distributions $\mathcal D^k = \mathcal{D}, \forall k\in[K]$, we call the samples from all clients are Independent and Identically Distributed (IID); otherwise, they are non-IID. We consider non-IID distribution like \cite{mcmahan2017communication}, where the label space on client $k$ is a subset of the total label space [$L$]. 
In real-world scenarios, label information from human annotators can be imperfect \cite{wei2022learning}, leading to noisy labels $\tilde{y}_{n}$ that may or may not be identical to the true labels $y_{n}$. If $\tilde{y}_{n} \neq {y}_{n}$, we say $\tilde{y}_{n}$ is corrupted, otherwise it is clean. Assume the label noise is class-dependent \cite{natarajan2013learning,liu2015classification,liu2020peer}. Then for each client $k$, we can use the noise transition matrix $T^{k}$ to capture the transition probability from clean label $Y=y$ to noisy label $\widetilde Y=\tilde y$. Specifically, we have the following definition: 
\begin{defn}[Client-Dependent Label Noise]\label{def:cdn}
The label noise on clients $\{1,\cdots, K\}$ is client-dependent if the label noise on each client $k$ can be characterized by noise transition matrix $T^k$, where each element satisfies:
\[\forall k\in[K], i,j\in[L], T^{k}_{ij}:=\PP_{(X,Y)\sim \mathcal D^k}(\widetilde Y=j | Y=i).\]
\end{defn}
When {\color{black}$\mathcal D^k = \mathcal{D}$ and} $T_k=T,\forall k\in[K]$, we refer to the label noise distribution among clients as homogeneous, otherwise, they are considered heterogeneous. This means that the heterogeneity of label noise distribution among clients primarily depends on the distribution of true labels and the associated label noise transition based on the true labels. {\color{black} For instance, if the noise transition matrix $T^k, k=1,.., K$ are heterogeneous, label noise distributions among clients are heterogeneous even if the distributions of true labels among clients are the same (as in the case of IID distribution).}

\section{\fedual:  Dual Model Structure Approach}  
\label{sec:4}


In this section, we introduce \fedual, the proposed dual structure approach to effectively train the FL model by identifying and filtering noisy label samples on clients. 



















\subsection{Dual Model Structure}
We present a novel dual model structure, illustrated in Fig. \ref{fig:figure2}, inspired by the well-known noisy label learning method Co-teaching \cite{han2018co}. The objective of \fedual\ is: 
\begin{equation}
  \min _{w}\left\{F(w) \triangleq \sum_{k=1}^{K} \frac{\bar{n}_{k}}{\bar{n}} F_{k}(w)\right\}
  \label{obj:F_w}
\end{equation}    

\noindent where $w$ is the parameters of the global model, $\frac{\bar{n}_{k}}{\bar{n}}$ is the weight of the $k$-th device, $\frac{\bar{n}_{k}}{\bar{n}} \geq 0$, and $\sum_{k}\bar{n}_{k} = \bar{n}$. Here, $\bar{n}_{k}$ is the number of selected samples on $k$-th client and $\bar{n}$ is the sum of selected samples on all clients. For client $k$, $F_{k}(\cdot)$ represents the local optimization objective, which measures the loss of predictions on local data. It is defined as follows:
\begin{footnotesize}
\begin{equation}
\label{obj:F_k}
F_{k}(w) = \frac{1}{\bar{n}_{k}}\sum_{n \in [n_{k}]} v_{n} \cdot \ell(x_{n}, \tilde{y}_{n} ; \theta_k) + \frac{\lambda}{2}\left\|\theta_k-w \right\|^2
\end{equation}    
\end{footnotesize}

\noindent where $v_{n} \in \left\{0, 1\right\}$ indicates whether example $n$ is clean ($v_{n} = 1$) or not ($v_{n} = 0$), and $\ell(\cdot)$ is the loss function. Towards the optimization objective, the training process of \fedual\ comprises three stages:
\begin{itemize}

\item Step-1: The selected clients download the global model $w$ from the server. The number of selected clients is determined by $C$, which is the fraction of a fixed set of $K$ clients in each round.
\item Step-2: At each client, the training process involves updating the dual models locally with the selected clean samples for $E$ epochs. 

\item Step-3: The locally-computed parameter updates $\nabla w_{t}$ of the global model $w$ will be uploaded to the server for aggregation.
\end{itemize}
There are dual models on each client for local updates of Step-2. On client $k$, the dual models are denoted the global model $w_k$ and personalized model $\theta_k$. The loss function on the dual models on the batch data $\tilde{b}$ is denoted as $\ell(\tilde{b})$, where the loss of the global model is $\ell(\tilde{b}; w_k)$, and the loss of personalized model is $\ell(\tilde{b}; \theta_k)$. When a batch $\tilde{b}$ arrives, the personalized model $\theta_k$ and global model $w_k$ in the ‘Filter Noisy Data’ module respectively select a small proportion of low loss samples out of the mini-batch $\tilde{b}$ by sample selector defined in Eq. (\ref{eq:sample_sieve}), denoted as $\bar{b}_{\theta_k}$ and $\bar{b}_{w_k}$. 
\begin{footnotesize}
\begin{equation}
\label{eq:sample_sieve} 
v_n=\mathds{1}\!\left(\!\ell_{CE}\left(f\left(x_n\right), \tilde{y}_n\right)-\frac{1}{L} \sum_{\tilde{y} \in[L]} \! \ell_{CE}\left(f\left(x_n\right), \tilde{y}\right)\!<\! 0 \!\right)
\end{equation}
\end{footnotesize}

\noindent where $f(\cdot)$ denotes the logistical output of a classifier, and $\ell_{CE}(\cdot)$ is the Cross Entropy (CE) loss. By employing the indicator function $\mathds{1}(\cdot)$, the sample selector determine $v_n = 1$ when $\ell_{CE}\left(f\left(x_n\right), \tilde{y}_n\right)-\frac{1}{L} \sum_{\tilde{y} \in[L]} \! \ell_{CE}\left(f\left(x_n\right), \tilde{y}\right) < 0$, otherwise $v_n = 0$. The sample selector is used to judge {\color{black}whether the CE loss of the sample deviates from the mean of the sample's prediction distribution}. The quality of selecting clean examples is guaranteed in Theorem \ref{theo}. 
\begin{theorem}
\label{theo}
The sample selector defined in (\ref{eq:sample_sieve}) ensures that clean examples ($x_n$, $\tilde{y}_n = y_n$) will not be identified as noisy label samples if the model $f(\cdot)$’s prediction on $x_n$ is better than a random guess.
\end{theorem}

\noindent Finally, the selected instances are fed into its peer model to generate the loss in the ‘Dual Model Update’ module, denoted as $\ell(\bar{b}_{\theta_k}; w_k)$ and  $\ell(\bar{b}_{w_k}; \theta_k)$ for alternate parameter updates. 




In the dual model structure, the global model $w$ and the personalized model $\theta$ inherently incorporate distinct different prior knowledge.  Through updates based on mutual insights, both models can identify and filter out mislabeled instances. The global model $w$ can be updated through local updates and global aggregation. On the contrary, if the personalized model $\theta$ is only updated solely based on the local client's data, it is prone to overfitting due to the limited amount of data available. This overfitting can hurt the overall performance of the dual models. Therefore, we introduce the regularizers to prevent the overfitting of dual models, as discussed in the following subsections.


\subsection{Regularizers}
\label{subse:regularizers}
\paragraph{Confidence Regularizer} Effectively selecting clean label samples is crucial. The commonly used Cross Entropy (CE) loss is inadequate for learning models with noisy label samples \cite{ghosh2017robust}. This could magnify the overfitting inclination of inherently susceptible personalized model in the dual models.
{\color{black}To address this, we introduce the Confidence Regularizer (CR) \cite{sieve2020}, defined in Eq. (\ref{eq:eq1}), as a modification to the vanilla CE loss. This incorporation aims to guide the model towards better fitting clean datasets.} 
\begin{footnotesize}
\begin{equation}
\label{eq:eq1}
\ell_{\mathrm{CR}}\left(f\left(x_{n}\right)\right):=-\beta \cdot \mathbb{E}_{\mathcal{D}_{\widetilde{Y} \mid \text{$\widetilde{D}$}}}\left[\ell_{CE}\left(f\left(x_{n}\right), \widetilde{Y}\right)\right]
\end{equation}    
\end{footnotesize}

\noindent Here, $\beta \geq 0$ represents {\color{black}a hyperparameter}, and the prior probability $\mathbb{P}(\widetilde{Y}\!\!\mid\!\!\widetilde{D})$ is determined based on the noisy dataset, {\color{black}i.e., $\mathbb P(\tilde Y=i) = \#\text{Label}\_i / N$ ($\#\text{Label}\_i$ is the number of samples for $i$-th class label, and N is the total number of samples)}. \xy{add some explanation}The incorporation of the CR leads to the following objective: 
\begin{footnotesize}
\begin{equation}
\begin{aligned}
\min\limits_{\substack{f \in \mathcal{F}, \\ v \in\{0,1\}^{n_k}}} &  \, \frac{1}{\bar{n}_k}\sum\limits_{n \in[n_k]} v_{n}\left[\ell_{CE}\left(f\left(x_{n}\!\right), \tilde{y}_{n}\right)+\ell_{CR}\left(f\left( x_{n} \right) \right) \right] \\
s.t.\quad\,\, & \ell_{CR}\left(f\left(x_{n}\right)\right):=-\beta \cdot \mathbb{E}_{\mathcal{D}_{\tilde{Y} \mid \widetilde{D}}}\ell\left(f\left(x_{n}\right), \widetilde{Y}\right) \\
\end{aligned}
\label{CR loss} 
\end{equation}    
\end{footnotesize}

\noindent The training of dual models is performed using the function $\ell(\cdot) = \ell_{CE}(\cdot) + \ell_{CR}(\cdot)$, and the sample selector defined in Eq. (\ref{eq:sample_sieve}) is used to differentiate between noisy and clean samples. Throughout the paper, we use $\ell(\cdot)$ to refer to $\ell_{CE}(\cdot) + \ell_{CR}(\cdot)$ by default. 
Consequently, the model can be updated on $(x_n,\bar{y}_n)$ according to Eq. (\ref{sieve_update}), where $(x_n,\bar{y}_n)$ is the sample selected by sample selector.
\begin{footnotesize}
\begin{equation}
    f =\underset{f \in \mathcal{F}}{\arg \min } \frac{1}{\bar{n}_k}\underset{n \in[\bar{n}_k]}{\sum} \ell_{CE}\left(f\left(x_n\right), \bar{y}_n\right)+\ell_{\mathrm{CR}}\left(f\left(x_n\right)\right)
    \label{sieve_update}
\end{equation}    
\end{footnotesize}

\noindent With the CR, the dual models' personalized model $\theta$ and global model $w$ can effectively, to a certain extent, avoid overfitting to noisy label samples. Subsequently, they update their parameters according to Eq. (\ref{sieve_update}) using the selected samples, as determined by the sample selector constructed by their peer model. However, it doesn't fundamentally address the issue of overfitting to local noisy data within the personalized model of the dual models. Consequently, the personalized model's updates drift farther away from the global model, causing the dual model structure to deteriorate. Therefore, we introduce a distance regularizer to effectively control this deviation.

\paragraph{Distance Regularizer} 
Instead of directly optimizing the objective (\ref{CR loss}), we can optimize the objective (\ref{obj:F_k}) by incorporating a Distance Regularizer (DR) into the loss function for each client. The DR can be defined as follows:
\begin{footnotesize}
\begin{equation}
 \frac{\lambda}{2}\left\|\theta_k-w \right\|^2
\label{eq:Fk}
\end{equation}    
\end{footnotesize}

\noindent where $\theta_k$ denotes the personalized model of client $k$, and $\lambda \in (0, +\infty)$ is a regularization parameter that governs the influence of the global model $w$ on the personalized model. A larger value of $\lambda$ enhances the benefits of the personalized model from the global model, while a smaller value encourages the personalized model to pay more attention to local information.
For solving objective (\ref{obj:F_k}), each client $k$ obtains its personalized model $\theta_k(w_k)$ through the update equation: 
\begin{footnotesize}
\begin{equation}
\theta_k(w_k) \gets \theta_k - \zeta (\bigtriangledown \ell_k\left(\theta_k\right) + \lambda(\theta_k - w_k))   
\end{equation}    
\end{footnotesize}

\noindent where $w_k$ is constant and refers to the local model of the client $k$, and $\zeta$, represents the personalized learning rate. Additionally, the global model of client $k$ is updated using: 
\begin{footnotesize}
\begin{equation}
\label{eq:w_update_DR}
w_k \gets w_k -\eta \lambda (w_k - \theta(w_k))
\end{equation}    
\end{footnotesize}

\noindent where $\eta$ denotes the global learning rate.

Please note that the previous updates of the global model in Eq. (\ref{eq:w_update_DR}) on clients are solely for solving the problem (\ref{eq:Fk}). However, it is crucial to emphasize that the global model should also be carried out for alternate updates, as illustrated in Algorithm \ref{alg:algorithm} (line 25). {\color{black}}

\begin{algorithm}[t]
\SetKwData{In}{\textbf{in}}\SetKwData{To}{to}
\DontPrintSemicolon
\SetAlgoLined
\KwIn { $T$, $T_s$, $E$, $\lambda$, $\eta$, $\zeta$, $\gamma$, $B$, $K$, $w$.}
\KwOut { $w$, $\theta_i, i=1,...,K$.}

\textbf{Server executes:}\\
initialize $w_{0}$\;
\For{each round $t$ from 0 to $T-1$}{
    $m \gets \max(C \cdot K, 1)$\;
    $\mathcal{S}_{t}$ $\gets$ (random set of $m$ clients)\;
    \For{each client $k \in \mathcal{S}_{t}$ \textbf{in parallel}}{
      $\bar{n}_{k}$, $w_{t+1}^{k} \gets$  \texttt{ClientUpdate}$(t, k, w_t)$\;}
    $w_{t+1} \gets (1-\gamma)w_t + \gamma \sum_{k \in |\mathcal{S}_{t}|}\frac{\bar{n}_{k}}{\bar{n}}w_{t+1}^{k}$\;}

\textbf{Client executes:} \algcom{/* Run on client $k$ */}\\
\texttt{ClientUpdate}($t$, $k$, $w$):\\
$\mathcal{\tilde{B}} \gets$ (split $\tilde{D}_{k}$ into batches of size $B$)\; 
$\theta \leftarrow \textit{deepcopy}(w)$\;
\For{each local epoch $i$ from 1 to $E$}{
    \For{batch $\tilde{b}^j \in \mathcal{\tilde{B}}$}{
     \eIf{$t \leq T_s$}
        {
           $\bar{b}_{w}^j = \bar{b}_{\theta}^j = \tilde{b}^j$\;
        }
        {
           $\bar{b}_{w}^j = Selector \, \ell_{CE}(\tilde{b}; w)$\;
           
           $\bar{b}_{\theta}^j = Selector \, \ell_{CE}(\tilde{b};\theta)$ \;
        }
     
     $\theta(w) \gets \theta - \zeta (\bigtriangledown \ell(\bar{b}_{w}^j;\theta) + \lambda(\theta - w))$\;
     $w \gets w -\eta \lambda (w - \theta(w))$\;
     
     $w \gets w -\eta \bigtriangledown \ell(\bar{b}_{\theta}^j; w)$\;}
     
     $\bar{n}_k^i = \sum_{j=1}^{B} |\bar{b}_{\theta}^j|$\;}
     
     $\bar{n}_{k} = \frac{1}{E}\sum_{i=1}^E \bar{n}_k^i $\;
     \Return{$\bar{n}_{k}$, $w$}
\caption{\emph{\fedual.}}
\label{alg:algorithm}
\end{algorithm}

\begin{table}[t]
\footnotesize
\centering
\scalebox{1}{\begin{tabular}{lllll}
\hline
\specialrule{0.05em}{3pt}{3pt} 

Dataset & Clothing1M & 
CIFAR-10 & MNIST  \\
\specialrule{0.05em}{3pt}{3pt}  

Training instances & 1,000,000 & 
50,000 & 60,000 \\
Number of classes & 14 &
10 & 10 \\
Total clients & 300 
& 100 & 100 \\
Fraction $C$ & 0.03 
& 0.1 & 0.1 \\
Learning rate $\eta$ & 0.001 
& 0.01 & 0.1\\
Rounds & 50 &
450 & 300 \\
Model Architecture & ResNet-50 & 
ResNet-18 & LetNet-5 \\
\hline
\end{tabular}}
\caption{Summary of the datasets. }
\label{tab:data1}
\end{table}

\begin{table*}[!t]
    \footnotesize
    \centering
    \scalebox{1}{
    \begin{tabular}{c|l|cccccc}
    \toprule\midrule
        \multicolumn{1}{c|}{\multirow{3}{*}{Datasets}} & \multicolumn{1}{c|}{\multirow{3}{*}{Methods}} & \multicolumn{3}{c}{\multirow{1}{*}{IID}} & \multicolumn{3}{c}{\multirow{1}{*}{non-IID}} \\
        \cmidrule(lr){3-5} \cmidrule(lr){6-8}
        &  & $\rho$ = 0.0 & $\rho$  = 0.5 & $\rho$  = 1 & $\rho$ = 0.0 & $\rho$ = 0.5 & $\rho$ = 1 \\ 
        &  & $\tau$ = 0.0 & $\tau$ = 0.3 & $\tau$ = 0.5 & $\tau$ = 0.0 & $\tau$ = 0.3 & $\tau$ = 0.5 \\ 
\midrule
\multirow{8}{*}{MNIST} & Local + CORES$^2$ & 96.79 $\pm$ 0.05 & 62.38 $\pm$ 3.88 & 37.30 $\pm$ 1.62 & 97.52 $\pm$ 0.17 & 90.31 $\pm$ 3.29 & 55.80 $\pm$ 6.45 \\
~ & Global + CORES$^2$ & 98.05 $\pm$ 0.05 & 97.39 $\pm$ 0.13 & 80.98 $\pm$ 4.78 & 97.46 $\pm$ 0.53 & 97.38 $\pm$ 0.12 & 87.45 $\pm$ 0.48 \\
~ & FedAvg & 98.39 $\pm$ 0.04 & 97.66 $\pm$ 0.09 & 93.90 $\pm$ 0.31 & 97.46 $\pm$ 0.57 & 97.73 $\pm$ 0.05 & 95.57 $\pm$ 0.28 \\
~ & FedProx & 98.22 $\pm$ 0.08 & 96.49 $\pm$ 0.08 & 93.90 $\pm$ 0.64 & 93.69 $\pm$ 4.98 & 97.25 $\pm$ 0.16 & 95.22 $\pm$ 0.37 \\

~ & RFL$^{\star}$ & 90.70 $\pm$ 0.54 & 96.54 $\pm$ 0.12 & 96.64 $\pm$ 0.08 & 90.51 $\pm$ 0.36 & 95.61 $\pm$ 0.34 & 91.56 $\pm$ 8.37 \\
~ & MR & 97.41 $\pm$ 0.21 & 95.98 $\pm$ 0.36 & 88.47 $\pm$ 1.48 & 97.09 $\pm$ 0.54 & 95.35 $\pm$ 0.54 & 90.53 $\pm$ 2.26 \\
~ & FedCorr$^{\star}$ & \textbf{98.68 $\pm$ 0.16} & \textbf{98.09 $\pm$ 0.22} & 95.67 $\pm$ 0.22 & 97.49 $\pm$ 0.82 & 97.75 $\pm$ 0.17 & 95.75 $\pm$ 0.46 \\\midrule
~ & \textbf{\fedual} & 98.07 $\pm$ 0.02 & 97.80 $\pm$ 0.18 & \textbf{96.79 $\pm$ 0.92}  & \textbf{98.05 $\pm$ 0.06} & \textbf{98.01 $\pm$ 0.18} & \textbf{96.00 $\pm$ 0.24} \\
\midrule
\multirow{8}{*}{CIFAR-10} & Local + CORES$^2$ & 84.23 $\pm$ 0.26 & 67.96 $\pm$ 0.60 & 22.51 $\pm$ 1.79 & 86.16 $\pm$ 0.71 & 68.46 $\pm$ 2.38 & 26.96 $\pm$ 1.21 \\
~ & Global + CORES$^2$ & 91.31 $\pm$ 0.09 & 85.81 $\pm$ 0.26 & 54.22 $\pm$ 3.18 & 90.27 $\pm$ 0.17 & 86.23 $\pm$ 0.20 & 35.31 $\pm$ 3.51 \\
~ & FedAvg & 90.33 $\pm$ 0.11 & 77.93 $\pm$ 0.29 & 33.87 $\pm$ 0.19 & 89.82 $\pm$ 0.34 & 78.30 $\pm$ 0.24 & 28.77 $\pm$ 0.59 \\
~ &  FedProx & 91.12 $\pm$ 0.20 & 79.33 $\pm$ 0.21 & 35.38 $\pm$ 0.31 & \textbf{90.50 $\pm$ 0.27} & 80.13 $\pm$ 0.40 & 29.63 $\pm$ 1.17 \\
~ &  RFL$^{\star}$ & 85.54 $\pm$ 0.26 & 84.06 $\pm$ 0.24 & 54.83 $\pm$ 1.06 & 83.83 $\pm$ 0.41 & 83.68 $\pm$ 0.32 & 49.49 $\pm$ 1.77 \\
~ &  MR & 70.65 $\pm$ 1.61 & 50.02 $\pm$ 2.13 & 23.67 $\pm$ 1.37 & 68.80 $\pm$ 0.52 & 51.46 $\pm$ 1.29 & 22.23 $\pm$ 1.72 \\
~ & FedCorr$^{\star}$ & \textbf{91.82 $\pm$ 0.22} & \textbf{88.05 $\pm$ 0.69} & 52.30 $\pm$ 0.91 & 81.07 $\pm$ 1.06 & 87.63 $\pm$ 0.64 & 45.43 $\pm$ 3.36 \\
\midrule
~ & \textbf{\fedual} & 90.72 $\pm$ 0.47 & 87.06 $\pm$ 0.30 & \textbf{62.87 $\pm$ 0.17} & 89.76 $\pm$ 0.32 & \textbf{87.82 $\pm$ 0.22} & \textbf{59.01 $\pm$ 0.55} \\

\bottomrule
\end{tabular}}
    \caption{Average (5 trials) accuracies (\%) of various methods on MNIST and CIFAR-10 datasets with IID and non-IID settings at different noise levels ($\rho$: ratio of noisy clients, $\tau$: lower bound of client noise level). The best results are highlighted in bold. Methods trained with label correction are marked by $\star$. }
\label{tab:result1}
\end{table*}
\begin{table*}[t]
\footnotesize
\centering
\scalebox{1}{
\begin{tabular}{lllllllll}
\hline\hline
Algorithm & Local + CORES$^2$ & Global + CORES$^2$ & FedAvg & FedProx & FedCorr$^{\star}$ & RFL$^{\star}$ & MR & \fedual\ \\
Acc & \multicolumn{1}{c}{53.32} & \multicolumn{1}{c}{65.18} & 66.75 & 66.76 & {\color{black}67.50} & 65.41 & {\color{black}50.01} & \textbf{70.52} \\
\hline
\end{tabular}}
\caption{Accuracies (\%) of various methods on Clothing1M with non-IID setting. }
\label{tab:clothing1m}
\end{table*}

\begin{table}[t]
\footnotesize
\centering
\scalebox{1}{
\begin{tabular}{c|c|cccc}
\toprule
\midrule
\multirow{2}{*}{Dataset} & \multirow{2}{*}{Method}  & \multicolumn{4}{c}{Sampling probability for each class.} \\
 &  & $p$ = 0.3 & $p$ = 0.5 &  $p$ = 0.7 &  $p$ = 0.9 \\ \midrule
\multirow{6}{*}{MNIST} & FedAvg & 92.20 & 95.45 & 95.53 & 95.14  \\
  & FedProx & 91.91 & 93.51 & 94.89 & 90.39  \\
  & RFL$^{\star}$ & 33.80 & 94.12 & 95.57 & 95.26  \\
  & MR & 80.72 & 87.21 & 92.75 & 90.60  \\
  & FedCorr$^{\star}$ & 93.84 & 95.15 & 95.77 & 95.73  \\
  & \textbf{\fedual} & \textbf{95.21} & \textbf{95.76} & \textbf{96.32} & \textbf{95.78}  \\
\midrule
\multirow{5}{*}{CIFAR-10} & FedAvg & 32.48 & 35.89 & 43.88 & 39.27  \\
  & FedProx & 34.50 & 37.00 & 45.28 & 40.39  \\
  & RFL$^{\star}$ & 45.04 & 58.03 & 66.93 & 66.91  \\
  & MR & 26.00 & 29.00 & 30.74 & 29.98  \\
  & FedCorr$^{\star}$ & 48.69 & 54.53 & 61.13 & 64.60  \\
  & \textbf{\fedual} & \textbf{51.19} & \textbf{62.86} & \textbf{71.54} & \textbf{72.46}  \\
  
\bottomrule
\end{tabular}}
\caption{Average (5 trials) accuracies (\%) of various methods on MNIST and CIFAR-10 datasets with different non-IID settings. The noise level is set to $\rho \!= \!1, \tau \!= \!0.5$ on the MNIST dataset and $\rho \!= \!1, \tau \!= \!0.3$ for the CIFAR-10 dataset.}
\label{tab:results2}
\end{table}
\section{Experiments}
\label{sec:5}

In this section, we begin by conducting experiments on benchmark datasets with different settings of heterogeneous label noise to comprehensively compare the performance of various methods. Next, we design experiments to evaluate the performance of different methods under increasing levels of label noise heterogeneity. Furthermore, we observe and compare the denoising performance of different methods across all clients to assess their denoising stability on each client. Lastly, we conduct an ablation study to evaluate the effectiveness of the individual components within the \fedual. 


\subsection{Experimental Setup}

\paragraph{Baselines} To compare \fedual\ with centralized noisy label learning methods applied to local clients (only trained on local data to filter noisy samples) and the server (trained by FedAvg to filter noisy samples), we consider a SOTA method: CORES$^2$ \cite{sieve2020}. \xy{this method should be proof that it's a good method} We refer to the centralized noisy label learning method applied to local clients as ``Local + CORES$^2$''. Similarly, we refer to the centralized method applied to the server as ``Global + CORES$^2$''. For reference, we also compare \fedual\ with two vanilla methods of FL: FedAvg \cite{mcmahan2017communication} and FedProx \cite{li2020federated}. Furthermore, we also comprehensively compare SOTA federated noisy label learning methods, such as FedCorr \cite{xu2022fedcorr}, RFL \cite{yang2022robust}, and MR \cite{tuor2021overcoming}.

\paragraph{Implementation Details} We evaluate different methods on MNIST \cite{lecun1998gradient}, CIFAR-10 \cite{krizhevsky2009learning}, 
and Clothing1M \cite{xiao2015learning} datasets with specific rounds, model architectures, total clients, and the fraction of selected clients for each dataset, as summarized in Tab. {\color{black}\ref{tab:data1}}. The performance of various methods is tested based on the data division and noise model used in  FedCorr \cite{xu2022fedcorr}, where $p \in (0, L]$ denotes the class sampling probability of clients, $\rho \in [0, 1]$ denotes the ratio of noisy clients, and $\tau \in [0, 1]$ denotes the lower bound for the noise level of a noisy client{\color{black}, which means that the noise level can be determined by sampling from
the uniform distribution $U(\tau, 1)$.} To ensure fairness in the comparison, we set common parameters for all the methods: local epoch $E$ = 5, batch size $B$ = 32, and the local solver SGD. For our \fedual\ method, we set $\beta = 2$ for CIFAR-10/MNIST, $\beta = 1$ for Clothing1M.
The learning rate $\zeta$ is set to $\frac{1}{2}\eta$ for MNIST dataset and $\zeta$ = $\eta$ for remaining datasets. Additionally, we set the beginning round of noisy sample filtering $T_s = 10$, hyperparameters $\gamma = 1$, and $\lambda = 15$ for all datasets. For other methods, we use the default parameters. 

We use test accuracy to evaluate the performance of the FL task and $F$-score to evaluate the performance of label error detection. $F$-score is calculated by $F = \frac{2 \times \texttt{Precision} \times \texttt{Recall}}{\texttt{Precision} + \texttt{Recall}}$. The precision and recall can be calculated as $\texttt{Precision} =\frac{\sum_{n \in[N]} \mathds{1}\left(v_{n}=0, \tilde{y}_{n} = y_{n}\right)}{\sum_{n \in[N]} \mathds{1}\left(v_{n}=0\right)}$, $\texttt{Recall} =\frac{\sum_{n \in[N]} \mathds{1}\left(v_{n}=0, \tilde{y}_{n} = y_{n}\right)}{\sum_{n \in[N]} \mathds{1}\left(\tilde{y}_{n} = y_{n}\right)}$, where $v_{n} = 1$ indicates that $\tilde{y}_{n}$ is detected as a corrupted label, and $v_{n} = 0$ if $\tilde{y}_{n}$ is detected to be clean \cite{sieve2020}. 



\begin{figure}[t]
\centering
\subfigure[IID $\rho$ = 0.5, $\tau$ = 0.3]{\includegraphics[width=0.48\linewidth]{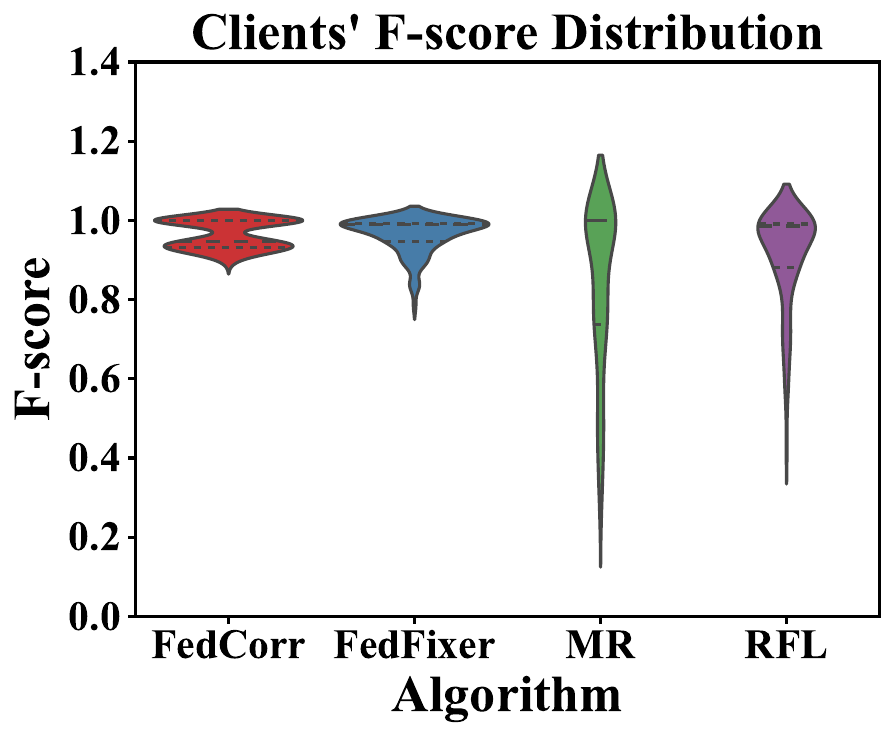}
}
\subfigure[IID $\rho$ = 1, $\tau$ = 0.5]{\includegraphics[width=0.48\linewidth]{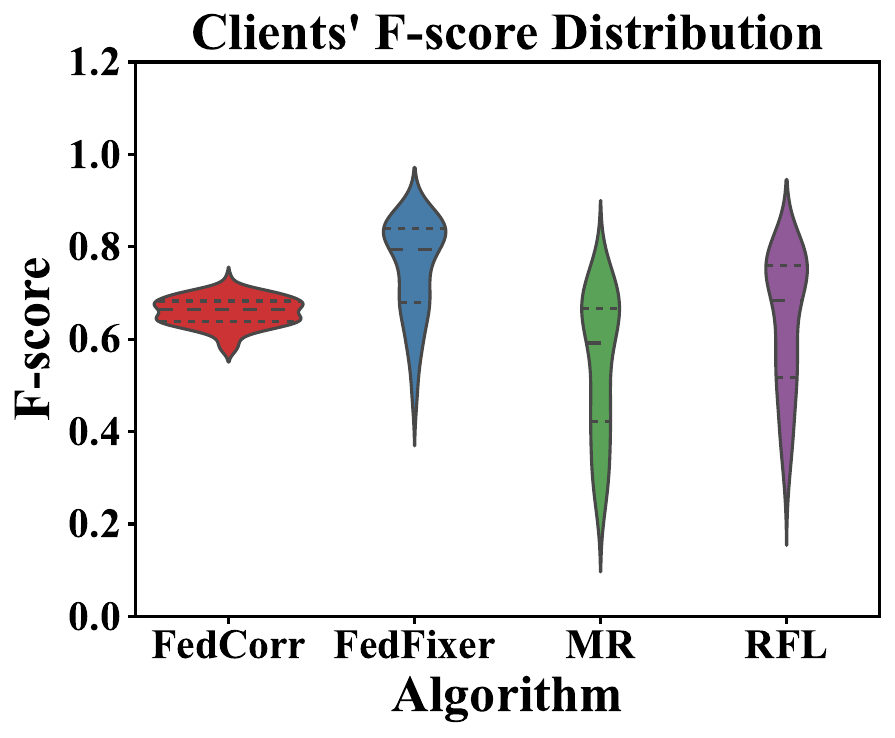}}
\subfigure[non-IID $\rho$ = 0.5, $\tau$ = 0.3]{\includegraphics[width=0.48\linewidth]{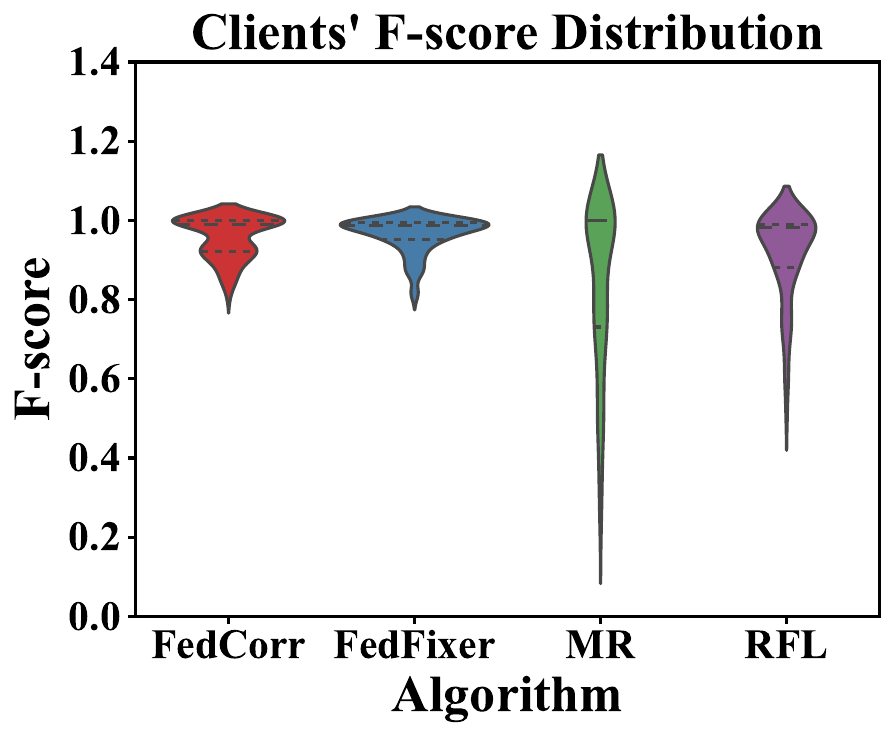}}
\subfigure[non-IID $\rho$ = 1, $\tau$ = 0.5]{\includegraphics[width=0.48\linewidth]{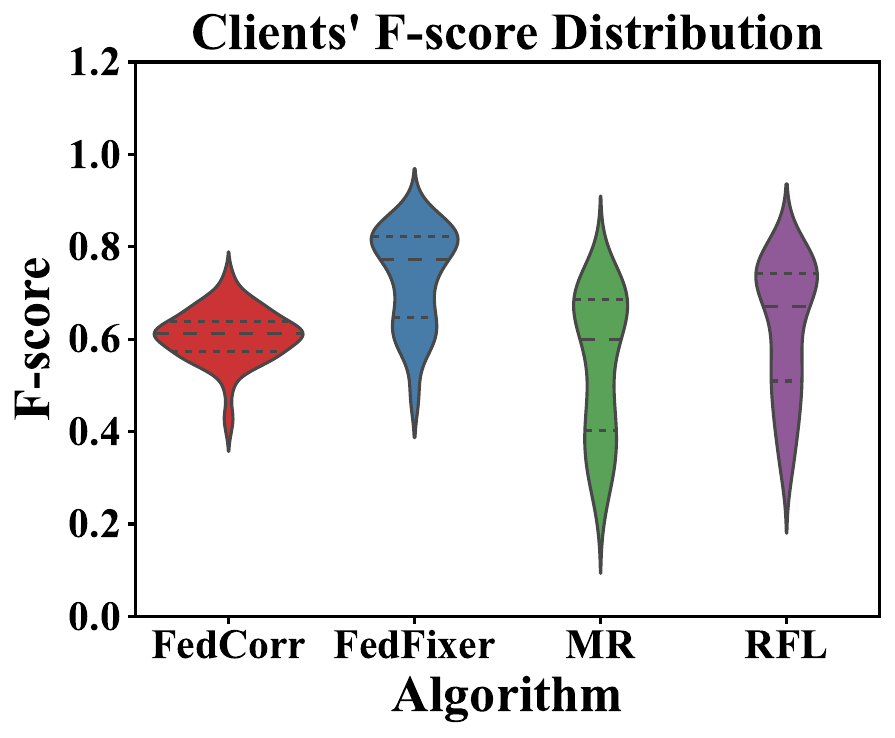}}
\caption{Clients' F-score distribution of different methods on CIFAR-10 with different noise levels on IID and non-IID distribution.}
\label{fig:3}
\end{figure}

\subsection{Comparison with State-of-the-Art Methods}
\label{Comparison}

\paragraph{Comprehensive Comparison} We compare the prediction performance of \fedual\ with all the baseline methods as mentioned above in heterogeneous label noise settings with both IID and non-IID distributions. The results are presented in Tab. \ref{tab:result1} for the MNIST, and CIFAR10
datasets. Firstly, we observe that \fedual\ consistently achieves excellent accuracy for heterogeneous label noise with IID or non-IID distribution at different noise levels. Secondly, in highly noisy settings, such as IID and non-IID with $\rho$ = 1 and $\tau$ = 0.5, \fedual\ significantly outperforms all other methods (especially on CIFAR-10). Furthermore, we notice that the method ``Local + CORES$^2$'' is almost ineffective and even collapses when directly applied to local clients to filter noisy label samples. Additionally, to assess compatibility, we also test the performance of \fedual\ on datasets without noisy label samples. The experimental results show that \fedual\ achieves a prediction accuracy similar to FedAvg and FedProx when the noisy label rate is 0.0. Tab. \ref{tab:clothing1m} shows the results on Clothing1M, a real dataset with noisy labels. For Clothing1M, \fedual\ achieves the highest accuracy.

\paragraph{Robustness with Heterogeneous Label Noise}

Tab. \ref{tab:results2} presents the performances of various methods as the heterogeneity of label noise increases. The degree of heterogeneity is determined by the sampling probability of each class, denoted as $p$, where $0 < p < 1$. This probability represents the likelihood of each class being present on each client. As the value of $p$ decreases, the level of label noise heterogeneity increases, creating a more challenging experimental scenario. In Tab. \ref{tab:results2}, the performances of all methods experience a decline with the increasing heterogeneity of label noise. Our method consistently achieves the highest prediction accuracies across all settings and notably outperforms others on the CIFAR-10 dataset. Different from the observations in Tab. \ref{tab:result1}, our experiments are always conducted under a high noise setting with $\rho = 1$, where the assumption is that all clients possess noisy labels. Under this scenario, the previously effective FedCorr method proves to be ineffective since it relies on the assumption of having a ratio of clean sample clients.
\paragraph{Denoising Stability}
\label{Denoising stability}




Fig. \ref{fig:3}  presents the F-score distribution of clients across different algorithms, evaluating the denoising stability on each client. In the violin plots, the distribution of \fedual\ and the FedCorr method appears to be highly concentrated, exhibiting consistently high mean values across all the other methods. When compared with FedCorr, our approach exhibits even denser distributions or higher mean values. Although in the IID scenario with $\rho$ = 1 and $\tau$ = 0.5, our approach exhibits a relatively scattered distribution, it still showcases a mean value nearly 0.2 higher than that of FedCorr. Consequently, it demonstrates approximately 10\% better predictive performance than FedCorr, as evident from the average test accuracy values in Tab. \ref{tab:result1} (IID, $\rho$ = 1, $\tau$ = 0.5). Similarly, in the IID scenario with $\rho$ = 0.5 and $\rho$ = 0.3, our method also exhibits a denser distribution of higher mean values. However, due to our initial approach not incorporating label correction, it shows a predictive accuracy lower by 1\% compared to FedCorr. This information can be further referenced in Tab. \ref{tab:result1} under the IID scenario with $\rho$ = 0.5 and $\tau$ = 0.3.

\begin{figure}[t]
    \centering
    \subfigure[Training without CR.]{\includegraphics[width=0.48\linewidth]{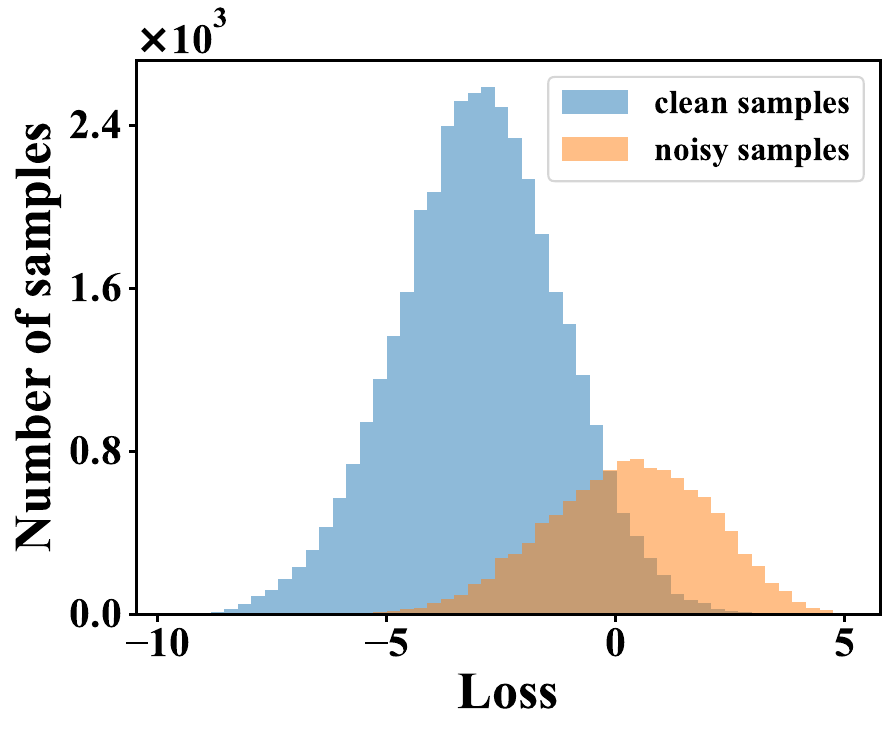}}
    \subfigure[Training with CR.]{\includegraphics[width=0.48\linewidth]{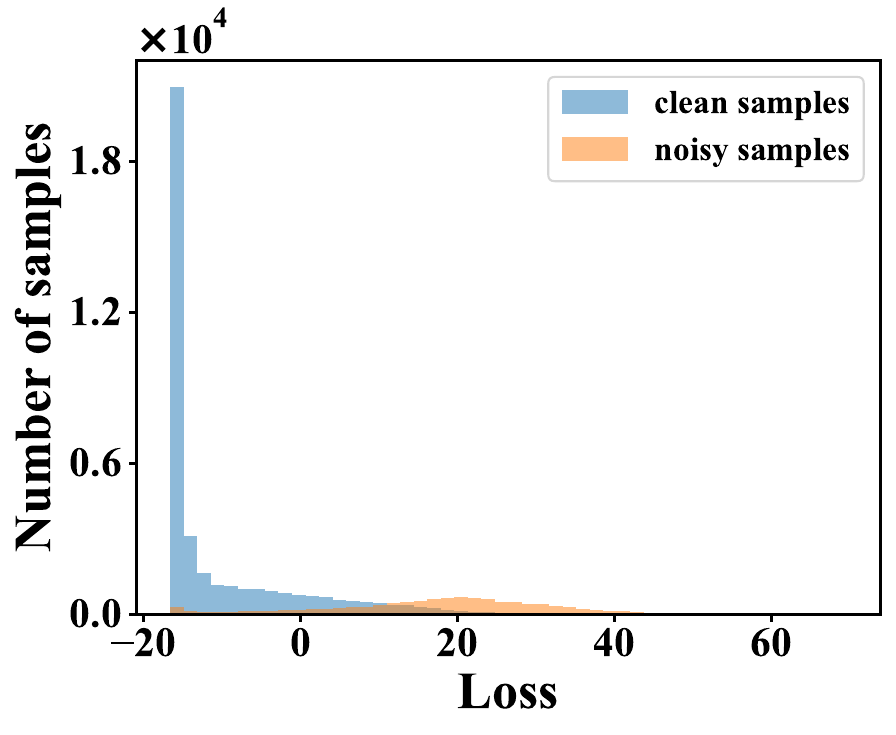}}
    
    \caption{Loss distribution of clean and noisy samples training without CR and with CR on CIFAR-10 with noise model $\rho = 0.5, \tau = 0.3$ and IID data partition (Round = 20). }
    \label{fig:enter-label}
\end{figure}
\begin{table}
    \footnotesize
    \centering
    \scalebox{1}{
    \begin{tabular}{l|cccc}
    \toprule
    \hline
    \multicolumn{1}{c|}{\multirow{3}{*}{Method}} &  \multicolumn{2}{c}{IID} & \multicolumn{2}{c}{non-IID} \\\cline{2-5}
    & $\rho$  = 0.5 & $\rho$  = 1.0 & $\rho$  = 0.5 & $\rho$  = 1.0 \\
    & $\tau$  = 0.3 & $\tau$  = 0.5 & $\tau$  = 0.3 & $\tau$  = 0.5 \\
    \midrule
    Ours & \textbf{87.06} & \textbf{62.87} & \textbf{87.82} & \textbf{59.01} \\
    Ours w/o DR & 84.45 & 39.62 &   83.94  &  35.74 \\
    Ours w/o CR & 75.73 & 35.50 &  76.29  & 30.56  \\
    Ours w/o AU & 86.10 &  62.62 & 87.30 & 54.84 \\
    Ours w/o PM & 85.47 &  50.10 & 85.92 & 34.44 \\
    \bottomrule
    \end{tabular}}
    \caption{Ablation study results on CIFAR-10.}
    \hspace{-1cm}
    \label{tab:ablation}
\end{table}

\subsection{Ablation Study}
{\color{black} Tab. \ref{tab:ablation} presents an analysis of the effects of different components within \fedual. This analysis involves evaluating the performance under various scenarios, including ours w/o CR (Confidence Regularizer), ours w/o DR (Distance Regularizer), ours w/o AU (Alternate Updates), and ours w/o PM (Personalized Model). As shown in Tab \ref{tab:ablation}, all these components contribute to performance improvement. The personalized model, in particular, plays a crucial role in scenarios with high levels of heterogeneous label noise. Moreover, Alternate Updates exhibit advantages in high levels of heterogeneous label noise characterized by non-IID distribution. In our dual models, CR has the largest effect. It improves the confidence predictions to effectively distinguish between clean and noisy samples, as depicted in Fig. \ref{fig:enter-label}. The DR is also a critical component, preventing the dual models from confusing noisy label samples with client-specific samples.}

\section{Conclusions}
\label{sec:6}
{\color{black}We introduce \fedual, a novel and effective dual model structure approach designed to address the issue of heterogeneous label noise in FL. In \fedual, a personalized model is introduced to learn client-specific samples to decrease the risk of misidentifying client-specific samples as noisy label samples. Moreover, the design of alternative updates of the global model and personalized model can effectively prevent error accumulation from a single model over time. The extensive experiments showcase the superior performance of \fedual\ across both IID and non-IID data distribution with label noise, particularly excelling in scenarios with high levels of heterogeneous label noise.}




\section*{Acknowledgements}
{\color{black}This work is supported by National Key R\&D Program of China under grants 2022YFF0901800, and the National Natural Science Foundation of China (NSFC) under grants NO. 61832008, 62072367, and 62176205. 
}

\bibliography{ref}

\end{document}